\documentclass[10pt,twocolumn,letterpaper]{article}

\usepackage{cvpr}              % To produce the CAMERA-READY version
\usepackage{algorithm}
\usepackage{algorithmic}
\usepackage{amsmath}
\usepackage{bbding}
\usepackage{booktabs}

\usepackage{array,multirow}
\usepackage{arydshln}
\usepackage[accsupp]{axessibility}
\usepackage{xcolor}
\usepackage{amsfonts}
\usepackage{graphicx}
\usepackage{mathtools}
\usepackage{url}
\usepackage{wrapfig}
\usepackage[export]{adjustbox}
\usepackage{bm} % bold math
\usepackage{caption}
\usepackage{subcaption}
\usepackage{amssymb}% http://ctan.org/pkg/amssymb
\usepackage{pifont}% http://ctan.org/pkg/pifont
\newcommand{\xmark}{\ding{55}}%
\usepackage[pagebackref,breaklinks,colorlinks]{hyperref}
\usepackage[capitalize]{cleveref}
\crefname{section}{Sec.}{Secs.}
\Crefname{section}{Section}{Sections}
\Crefname{table}{Table}{Tables}
\crefname{table}{Tab.}{Tabs.}

\begin{document}

\title{Motion-inductive Self-supervised Object Discovery in Videos}
\author {
    % Authors
    Shuangrui Ding\textsuperscript{\rm 1} \quad
    Weidi Xie\textsuperscript{\rm 1} \quad Yabo Chen\textsuperscript{\rm 1}  \\ Rui Qian\textsuperscript{\rm 2} \quad Xiaopeng  Zhang\textsuperscript{\rm 3}  \quad Hongkai Xiong\textsuperscript{\rm 1} \quad Qi Tian\textsuperscript{\rm 3} \\
    % \thanks{ Corresponding author. Email: xionghongkai@sjtu.edu.cn.}
    \textsuperscript{\rm 1}Shanghai Jiao Tong University \quad \textsuperscript{\rm 2}The Chinese University of Hong Kong \quad \textsuperscript{\rm 3}Huawei Inc. \\
    {\tt\small \{dsr1212, weidi, chenyabo, xionghongkai\}@sjtu.edu.cn}\\
    {\tt\small qr021@ie.cuhk.edu.hk  zxphistory@gmail.com tian.qi1@huawei.com}
}

\maketitle

\begin{abstract}
In this paper, we consider the task of unsupervised object discovery in videos. Previous works have shown promising results via processing optical flows to segment objects. However, taking flow as input brings about two drawbacks. First, flow cannot capture sufficient cues when objects remain static or partially occluded. Second, it is challenging to establish temporal coherency from flow-only input, due to the missing texture information. 
To tackle these limitations, we propose a model for directly processing consecutive RGB frames, and infer the optical flow between any pair of frames using a layered representation, with the opacity channels being treated as the segmentation. Additionally, to enforce object permanence, 
we apply temporal consistency loss on the inferred masks from randomly-paired frames, which refer to the motions at different paces, and encourage the model to segment the objects even if they may not move at the current time point.
Experimentally, we demonstrate superior performance over previous state-of-the-art methods on three public video segmentation datasets (DAVIS2016, SegTrackv2, and FBMS-59), while being computationally efficient by avoiding the overhead of computing optical flow as input.
\end{abstract}
\begin{figure*}
    \centering
    \includegraphics[width=\textwidth]{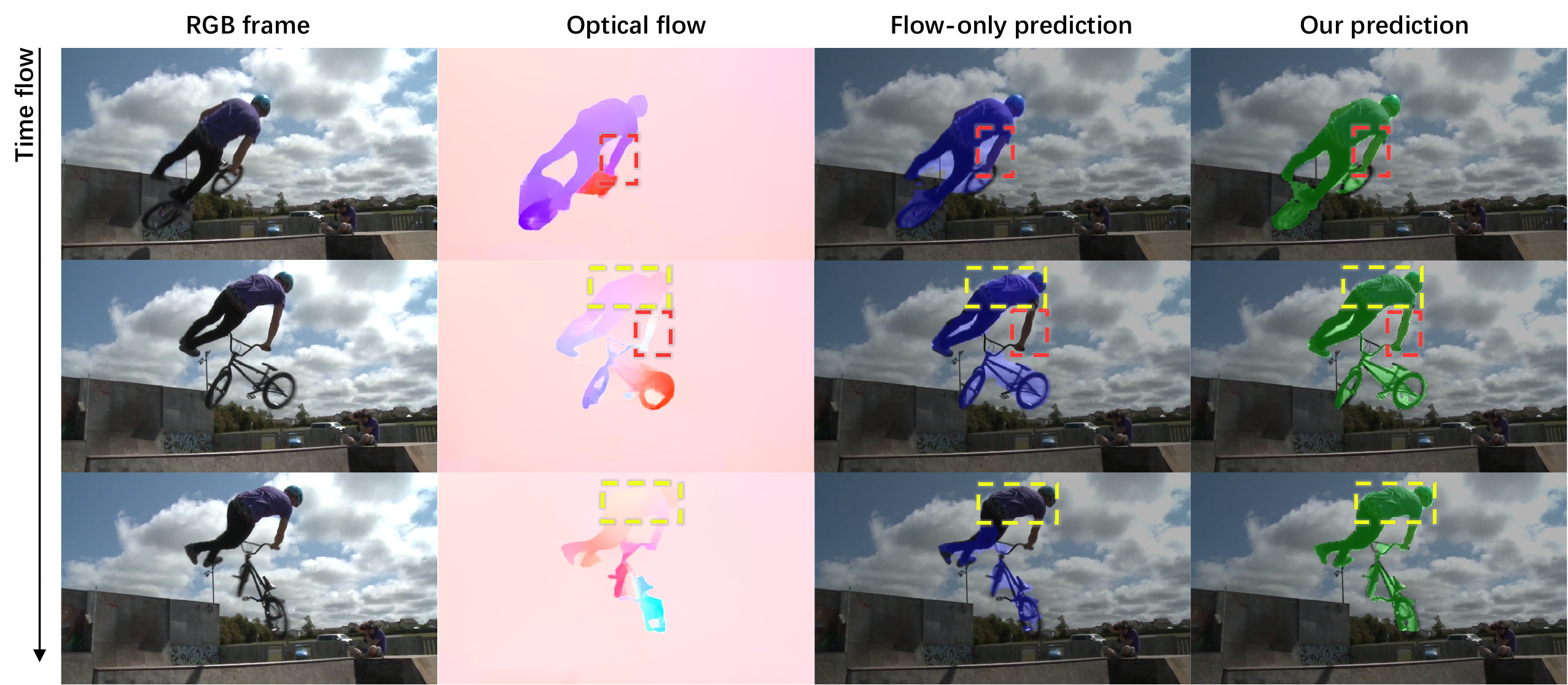}
    \caption{Illustration about Bicycle Motocross (BMX) sequence on SegTrackv2~\cite{li2013video}. The red boxes and the yellow boxes refer to the arm and the back of the player, respectively. Flow-only method~\cite{Yang_2021_ICCV} fails to track the same region in a temporal consistent fashion since it derives the foreground region directly from current optical flow. However, our methodology of processing a {\em RGB} video clip develops a sense of object permanence and solves the issue. }
    \label{fig:teaser}
\end{figure*}
\section{Introduction}

% Most animals including human beings are born to perceive the visual world thanks to the highly evolved biological visual system. 
% One exceptional function is to segment moving objects inherently~\cite{johansson1975visual}. Relying on the principle of common fate~\cite{spelke1990principles}, even infants are capable of discriminating the object that moves as a whole and separating each other. Moreover, after memorizing and grasping the shape and color of a certain moving object, the visual system could further discover the object even if it is motionless. In other words, objectness can be learned by temporal dynamics and appearance invariance without prior semantic knowledge. This cognitive feature lays the foundation for creatures to understand and interact with complex scenes. Motivated by this nature, the community of computer vision explores whether the learning-based models can tell the object by video data itself without any manual annotation~\cite{Yang_2019_CVPR, Yang_2021_CVPR, Yang_2021_ICCV, liu2021emergence, choudhury2022guess}. 

% \weidi{This is Weidi's version for 1st paragraph.}

Representing the visual scene with objects as the basic elements has long been considered a fundamental cognitive ability of the intelligent agent, for it enables understanding and interaction with the world more efficiently, for example, combinatorial generalization in novel settings~\cite{tenenbaum2011grow}. 
Although it remains somewhat obscure at the level of neurophysiology on exactly how humans discover the objects in a visual scene in the first place, it is a consensus that motion seems to play an indispensable role in defining and discovering the objects from the scene.
For example, in 1923, Wertheimer introduced the common fate principle that elements moving together tends to be perceived as a group~\cite{wertheimer1923untersuchungen}; while later Gibson claimed the independent motion has even been treated as one attribute to define an object visually~\cite{gibson1966senses}. 
Grounded on the above assumptions, the recent literature has witnessed numerous works with different models proposed for segmenting the moving objects via unsupervised learning~\cite{Yang_2019_CVPR, Yang_2021_CVPR, Yang_2021_ICCV, liu2021emergence}.

%As a consequence, apart from positing the existence of objects, 
%an ideal object-centric visual representation is also required to perceive the permanence over time, {\em i.e.}, realising that objects continue to exist even when they do not move explicitly.

Exploiting optical flows for object discovery naturally incurs two critical limitations: {\em First}, objects in videos may stop moving or be partially occluded at any time point, leaving no effective cues for their existence in the flow field; {\em Second}, computing optical flow from a pair of frames refers to a lossy encoding procedure, that poses a significant challenge for establishing temporal coherence, due to the lack of effective texture information. 
In contrast, adopting RGB frame sequences poses a few clear advantages. 
The most obvious one is that, while objects do not necessarily move all the time,
the property of temporal coherence in RGB space naturally guarantees a preliminary understanding of object permanence;
Additionally, the rich textures in the appearance stream give more distinctive patterns than those in motion, allowing to better identify and distinguish the different objects. Last but not least, processing RGB streams still enables a faster processing speed than using optical flow.

In this paper, our goal is to train a video segmentation model that can discover the moving objects within a sequence of RGB frames, in the form of segmentation. In specific, our proposed model first encodes consecutive frames independently, into a set of frame-wise visual features, 
that is followed by a temporal fusion with a Transformer encoder. 
To localise the moving objects, we randomly pair the visual features from two frames and pass them into a frame comparator module, 
effectively establishing the relative motion between frames.
Inspired by~\cite{Yang_2021_ICCV}, we decode the motion features into optical flows with a dual-layered representation, with the opacity weight of each layer treated as the segmentation mask. At training time, we exploit an off-the-shelf optical flow estimator, {\em e.g.}, RAFT~\cite{teed2020raft}, as the induction for flow reconstruction. To develop the property of object permanence, we enforce a temporal consistency on the inferred segmentation masks, which encourages the model to mine effective texture information from the RGB sequence and keep track of the objects even if they may be static at the current time point.

In short, we summarize the contributions in this paper: 
{\em First}, we introduce the \textbf{M}otion-inductive \textbf{O}bject \textbf{D}iscovery (\textbf{MOD}) model, a simple architecture for discovering the moving objects in videos, by directly processing a set of consecutive RGB frames.
{\em Second}, we propose a self-supervised proxy task that is used to train the architecture without relying upon any manual annotation.
To overcome the challenge from flow-based methods, {\em i.e.}, objects may stay static or move slowly, we adopt a random-paired policy and restrain the temporal consistency.
{\em Third}, we conduct a series of ablation studies to validate each key component of our method, such as the temporal consistency of random-paired flow. 
While evaluating three public benchmarks,
we demonstrate superior performance over existing approaches on DAVIS2016~\cite{perazzi2016benchmark}, SegTrackv2~\cite{li2013video}, and FBMS-59~\cite{ochs2013segmentation}, with considerable speed-up during the inference procedure.

\section{Motion-inductive Object Discovery Model}
\label{method}
\begin{figure*}
    \centering
    \includegraphics[width=\textwidth]{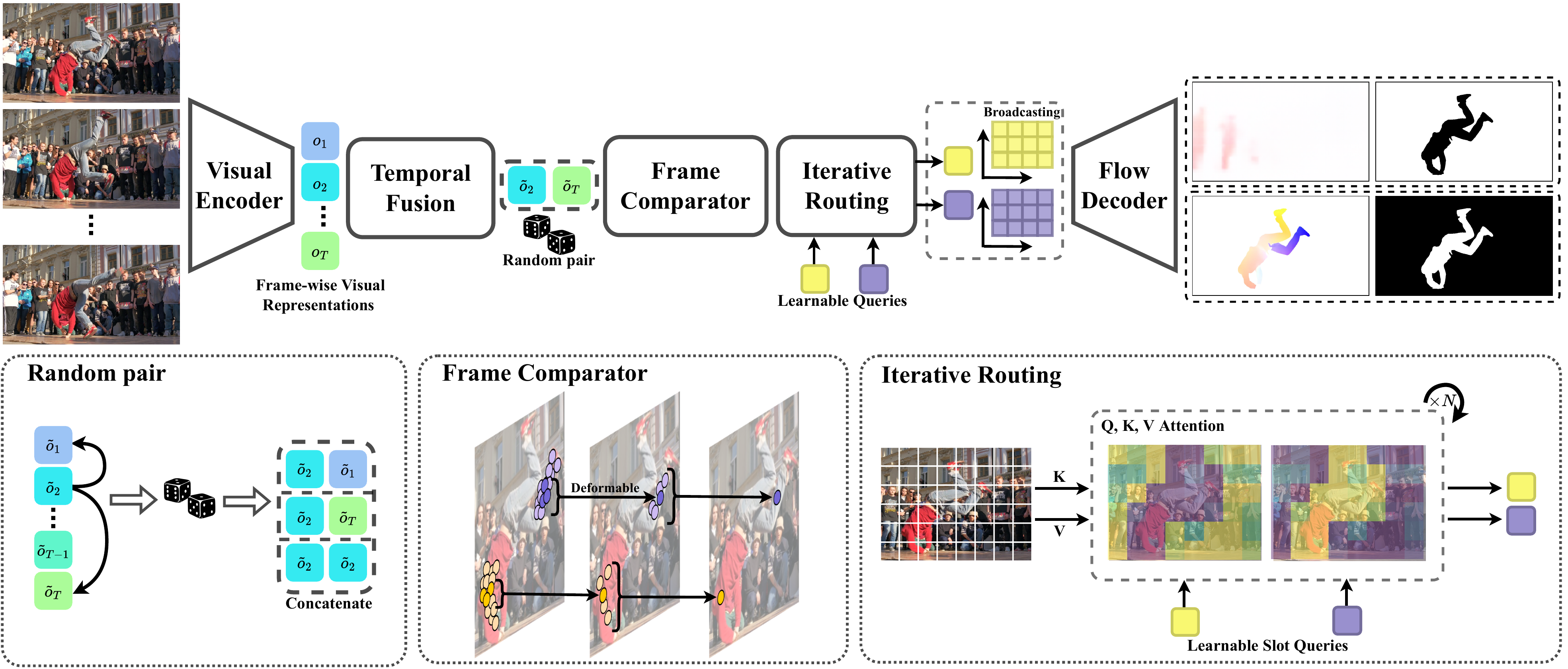}
    \caption{\textbf{Model architecture}. Our model first extracts spatial features of consecutive frames by the visual encoder. To jointly model the temporal relation, we aggregate and interact among the multi-frame features with late-fusion. We randomly pair two frames' visual representations and pass them into the frame comparator to encode relative motion. 
    Then, we decode the flow from random-paired frames. 
    Through iterative routing, we adopt a dual-layer representation for the flow reconstruction,
    {\em i.e.}, outputting the foreground and background flows separately, and composing them with inferred opacity weights. The whole training procedure does not require supervision from any mask annotations.}
    % \weidi{the figure looks too small, maybe consider to somehow zoom in specific module ?}
    % \dsr{I will figure out tomorrow.}
    \label{fig:pipeline}
\end{figure*}

In this section, we detail our Motion-inductive Object Discovery (MOD) model, which processes a set of consecutive RGB frames and automatically discovers the moving objects in the form of segmentation.
An overview of the training procedure can be seen in Figure~\ref{fig:pipeline}, where the visual features computed from individual frames are temporally fused, and randomly paired together, for decoding the optical flows between two corresponding frames. 
Taking inspiration from the motion grouping~\cite{Yang_2021_ICCV}, we also adopt a dual-layered representation for the output flow, 
with the foreground and background flows being reconstructed separately, and later composited with the inferred opacity masks~(soft segmentation).
In the following section, we will detail each key component in our proposed architecture.
%unlike previous works that process optical flows for individual frame, our model takes a short RGB video clip as input, {\em i.e.}~$V = \{X_1, \dots, X_T\}$, $ X_t \in \displaystyle \R^{H \times W \times 3}$, and learns to segment the moving objects in video clips by jointly reconstructing the optical flow between each pair of the frames.

% Note that, 

% In Section~\ref{sec:st_enc}, we start by detailing the proposed architecture and training procedure by flow reconstruction;
% to enforce the temporal coherence, in Section~\ref{sec:cons}
% we further introduce an auxiliary loss that enables to discover the salient objects even in static frames with the guidance of motion cues. 
% Finally, we include a discussion on our method in Section~\ref{sec:dis}.  

\subsection{Spatial-temporal Visual Encoder}
\label{sec:mf_enc}
To start with, our model takes a short video clip as input, {\em i.e.}, $v = \{x_1, \cdots, x_T\}, v \in \mathbb{R}^{T \times H \times W \times 3}$,
consisting of a set of RGB frames,
the frame-wise visual representations are computed with a shared visual encoder $\Phi_{\text{enc}}$. Formally, the output feature map $o_{t}$ at timestamp $t$ $( 1 \leq t \leq T)$ can be obtained:
\begin{equation}
o_{t} = \Phi_{\text{enc}}(x_t) \in   \mathbb{R}^{ h \times w \times d},  
\end{equation}
where $h, w, d$ denotes the dimension of the height, width, and channel, respectively.
%we can use any latest variants of visual backbones, e.g., CNNs and Transformers. 
% In this paper, we adopt the first three stages of SwinV2-Tiny~\cite{liu2022swin} as the frame encoder. 
Till this point, to build the temporal dependency between multiple visual frames,
a global fusion module is introduced, via a standard Transformer Encoder layer:
\begin{align}
\{\Tilde{o}_1, \dots, \Tilde{o}_T\} = \Phi_{\text{temp}}(\{o_1+\texttt{pe}_1, \cdots, o_T+\texttt{pe}_T\}),
\end{align}
where \texttt{pe} refers to the learnable spatial-temporal positional encodings,
% \weidi{is this learnable ? or sinusoidal ?} % DSR: learnable 
and the output $\Tilde{o}_t \in \mathbb{R}^{h \times w \times d}$ remains the same dimension as input. 

By taking the multiple RGB frames as input, our proposed visual encoder can explicitly consider the temporal coherence within the video clip. 
\textbf{Note that}, such seemingly simple design poses two critical differences from previous work on motion-driven object discovery~\cite{Yang_2019_CVPR, Yang_2021_ICCV}, where only frame-wise optical flow is adopted:
{\em First}, using RGB frames as input can drastically reduce the computation latency at inference time. The throughput of our model without computing dense optical flow reaches round \textbf{100 fps} on a standard 32GB Tesla V100 GPU while prior works need to calculate flow at first. RGB provides more semantic information than flows for the model to exploit, {\em i.e.}, including not only the object's shape but also its texture. 
% \weidi{add 1-2 sentences};
{\em Second}, processing multiple frames contributes to the development of a sense of object permanence within the video clip, {\em i.e.}, the understanding that items or people still exist even when they cannot be perceived explicitly. 
Therefore, even though the objects in videos may stop moving or be partially occluded at any time point, they can still be effectively segmented with the temporal cues.

\subsection{Random-paired Frame Comparator}
\label{flow-enc}

Till this point, we consider building the relative motion between any two visual frames within the video clip, from reference frame $i$ to target frame $j$, 
{\em i.e.}, $f_{i \rightarrow j}$. 
In specific, we select the visual representation of the two corresponding frames, {\em i.e.}, $\Tilde{o}_{i}$ and $\Tilde{o}_{j}$, 
concatenate them along the feature dimension, and feed it into a comparator module:
\begin{equation}
    f_{i\rightarrow j} = \Phi_{\text{comp}}(\text{concat}(\Tilde{o}_{i}, \Tilde{o}_{j})) \in  \mathbb{R}^{h\times w \times d},
\end{equation}
where $\Phi_{\text{comp}}:\mathbb{R}^{2d} \rightarrow \mathbb{R}^{d}$ consists of multiple deformable convolutional layers~\cite{zhu2019deformable} followed by a series of Transformer Encoders, that dynamically construct the feature representation for later estimating relative motion between frames while reducing the feature dimensions at the same time. 

\paragraph{Discussion.}
With such a design of random-paired frame comparator, 
our MOD is capable of modeling the relative motion between any two randomly sampled frames out of the video clip~($T$ frames), accounting for a total of $T^2$ frame pairs for forward, backward, single, or multi-step motions. Note that, we do not distinguish the case of $i \neq j$ and $i = j$. Specifically, the former encourages the model to discover the relative motion between two different frames, while the latter refers to an extreme case that neither objects nor camera is moving, and no motion cues are available, thus enforcing the model to discover objects via temporal coherence, {\em i.e.}, objects that moves in any frame along the video should also be discovered in static frames.

\subsection{Dual-layered Flow Decoder}
\label{flow-dec}
To decode the features into the form of optical flow,
we adopt several slot attention layers~\cite{locatello2020object} with two learnable queries, {\em i.e.}, termed as slot vectors, iteratively attending the output visual features from the comparator, and decoded into the optical flow between any two frames with a dual-layered representation. 
In detail, a slot attention module acts similarly to a Transformer Decoder,
with the only exception being that the normalisation is computed along the slot side,
thus each slot competes to take over the pixels. In each iteration, given two slot vectors as $S \in \mathbb{R}^{2 \times d}$ and visual feature maps $\Tilde{o}_t \in \mathbb{R}^{hw \times d}$, we use three linear projections to compute the \texttt{query}, \texttt{key} and \texttt{value}, {\em i.e.}, $Q \in \mathbb{R}^{2\times d}, K, V \in \mathbb{R}^{hw\times d}$.
Thereafter, we can obtain the weights matrix $W \in \mathbb{R}^{2 \times hw}$ and normalise along the slot dimension,  {\em i.e.}, 
% Since two slots are designed to compete for every pixel, the softmax function is utilized to normalize along slots dimension
\begin{equation}
% \begin{array}{cc}
\begin{aligned}
    &\widetilde{W}_{s,\cdot}  = \exp(W_{s,\cdot}) / \sum_{l}\exp{(W_{l,\cdot})}, \\
    &\text{ where } W = \frac{1}{\sqrt{d}}Q  K^{T}.  
\end{aligned}
\end{equation}

Then, we gain the next iteration's slot vectors by aggregating the values $V$ and passing them into a Gated Recurrent Unit (GRU)~\cite{cho2014learning}, {\em i.e.},
\begin{equation}
\begin{aligned}
    &S \coloneqq \text{GRU}(\text{inputs} = AV, \text{states}=S), \\
    &\text{ where }   A_{\cdot, s} = \frac{\widetilde{W}_{\cdot, s}}{\sum_{l}\widetilde{W}_{\cdot, l}} \in \mathbb{R}^{2 \times (h\times w)}.
\end{aligned}
\end{equation}
We iterate the whole routing process for $N$ times. In this way, the entities of similar RGB and flow patterns are grouped together and distinct pixels are separated by two slots. 
% Finally, we gain the next iteration's slots $S_{n+1}$ via a Gated Recurrent Unit~\cite{cho2014learning}. 
Eventually, we broadcast the final outputted slots into $G = \{G^s \in \mathbb{R}^{h\times w \times d}\}_{s=1}^{2}$ added with learnable spatial positional embeddings to construct the optical flow. A flow decoder $\Phi_{\text{dec}}$ consisting Transformer encoders and up-sampling layers % \weidi{convolutional and upsampling layers ?} 
takes the slot grids as input, and outputs dual layers of optical flows $\{\widetilde{I}^s \in \mathbb{R}^{H \times W \times 3}\}_{s=1}^{2}$~\footnote{Note that following the common practise, we also output 3-channel RGB images, that refers to a transformation from the traditional 2-channel optical flow based on the color wheel proposed in~\cite{sun2018pwc}.} and their opacity weights $\{\alpha^s \in \mathbb{R}^{H \times W \times 1}\}_{s=1}^{2}$: 
\begin{equation}
    \{\widetilde{I}^s, \alpha^s\}_{s=1}^{2} =  \Phi_{\text{dec}}(G),
\end{equation}
where $\alpha^s \in [0, 1]^{H \times W \times 1}$ is normalized across two slots via softmax function. 
For a given frame pair $(i,j)$, their relative flow $\widetilde{I}_{i\rightarrow j}$ can be computed via:
\begin{equation}
    \widetilde{I}_{i\rightarrow j} = \sum_{s=1}^2 \alpha^{s}_{i\rightarrow j} \otimes  \widetilde{I}^{s}_{i\rightarrow j},
\end{equation}
where $\otimes$ denotes the element-wise multiplication. At inference time, we adopt the binarized opacity weights $\alpha^{s}$ as the object segmentation masks.

\subsection{Training}
% For the fair comparison with MG~\cite{Yang_2021_ICCV}, we impose the similar loss combo on the training.  
In this section, we describe the training procedure for the proposed model, 
on the raw videos without using manual annotations for the object segmentations.
In general, the training loss is composed of three components, 
namely, flow reconstruction, temporal consistency, and entropy minimisation.

\paragraph{Flow Reconstruction.}
As the main objective for optimisation, we use the flow reconstruction, 
where we adopt an off-the-shelf optical flow estimator, for example, RAFT~\cite{teed2020raft}, to estimate the flow between any two frames in the video.
We minimise the discrepancy between the dual-layer flow reconstruction and the output from the existing flow estimator:
\begin{equation}
    \mathcal{L}_{\text{recon}}^{i\rightarrow j} = \frac{1}{|\Omega|} \sum_{u \in \Omega} |I_{i\rightarrow j}(u) - \widetilde{I}_{i\rightarrow j}(u)|_2,
\end{equation}
where $\Omega = \{1,\cdots, H\} \times \{1, \cdots, W\}$ represents the spatial lattice, and a $|\cdot|_2$ denotes the L2 norm. In practice, the reconstruction constraint for zero flow when $i=j$ is dropped due to the distribution gap between dynamic flow and static flow, resulting in the divergence of the training.
\textbf{Note that}, flow estimator is only used for model training,
at inference time, our proposed architecture directly processes the RGB video clips.

\paragraph{Temporal Consistency.}
In order to build up temporal consistency within the input video, 
the pair of motion embeddings $f_{i\rightarrow j}, f_{i\rightarrow k}$ that starts from the same reference frame $i$, are passed through the flow decoder to reconstruct the optical flow between two corresponding frames.
Note that, as $1 \leq j, k\leq T$ are randomly sampled at every training iteration,
the output flow will refer to the motion at a different pace.
However, the predicted alpha weights for flow composition denote the soft segmentation for the same objects in the $i$-th frame, thus remaining consistent. 
% \weidi{subject to a permutation ?}
In specific, the two inferred masks $\{\alpha_{i\rightarrow j}^s\}_{s=1}^{2}, \{\alpha_{i\rightarrow k}^s\}_{s=1}^{2}$ are enforced to pull closer by minimising mean-squared error $\mathcal{L}_{\text{cons}}$, {\em i.e.},
\begin{equation}    
\begin{aligned}
    \mathcal{L}_{\text{cons}} = \frac{1}{T} \sum_{i=1}^{T} \mathcal{L}_{\text{cons}}^i&, \\ \text{ where } \mathcal{L}_{\text{cons}}^{i} =   \frac{1}{2|\Omega|}\sum_{u \in \Omega} \sum_{s=1}^2|\alpha_{i\rightarrow j}^s(u) & -\alpha_{i\rightarrow k}^s(u)|^2.
\end{aligned}
\end{equation}

%In practice, we find our temporal consistency does not require the permutation-invariant design~\cite{Yang_2021_ICCV}, that is, for the same starting frame, the slot correspondence stays immutable.    

\paragraph{Entropy Minimisation.}
Lastly, we impose a pixel-wise entropy regularisation on inferred masks,
that is zero if the alpha channels are one-hot, and maximum when they are of equal probability. Intuitively, this helps encourage the masks to be binary, which aligns with our goal in obtaining segmentation masks:
\begin{equation}
    \mathcal{L}_{\text{entro}}^{i\rightarrow j} = \frac{1}{2|\Omega|}\sum_{u \in \Omega} \sum_{s=1}^2 -\alpha_{i\rightarrow j}^s(u) \log(\alpha_{i\rightarrow j}^s(u)).
\end{equation}

\paragraph{Total Loss.}
Accordingly, we rewrite aforementioned reconstruction loss $\mathcal{L}_{\text{recon}}$ and entropy regulatization $\mathcal{L}_{\text{entro}}$ in summed version:
\begin{align}
    \mathcal{L}_{\text{recon}} = \frac{1}{2T}\sum_{i=1}^{T} \mathcal{L}_{\text{recon}}^{i\rightarrow j_i} + \mathcal{L}_{\text{recon}}^{i\rightarrow k_i}; \\
    \mathcal{L}_{\text{entro}} = \frac{1}{2T}\sum_{i=1}^{T} \mathcal{L}_{\text{entro}}^{i\rightarrow j_i} + \mathcal{L}_{\text{entro}}^{i\rightarrow k_i}.    
\end{align}
The total loss for training our model can thus be computed as:
\begin{equation}
    \mathcal{L}_{\text{tot}} = 
    \lambda_{\text{r}} \mathcal{L}_{\text{recon}} +
    \lambda_{\text{e}} \mathcal{L}_{\text{entro}} + 
    \lambda_{\text{c}} \mathcal{L_{\text{cons}}},
\end{equation}
where we set $\lambda_{\text{r}}=100$, $\lambda_{\text{e}}=\lambda_{\text{c}}=0.01$ at the beginning of the training. 
We notice the model to be fairly robust to these hyper-parameters.
% \dsr{It should be in similar situation with MG.}

% \label{sec:dis}
\section{Experimental Setup}
In the experiments setup, we first introduce the benchmarks and then elaborate on implementation details. 
\subsection{Datasets}
\label{data}
We benchmark on three popular datasets designed for video object segmentation. 
\textbf{DAVIS2016}~\cite{perazzi2016benchmark} consists of 50 high quality videos, 3455 frames in total. Every frame is annotated with a pixel-accurate segmentation mask. 
\textbf{SegTrackv2}~\cite{li2013video} contains 14 sequences and 947 fully-annotated frames. Each sequence involves 1-6 moving objects and presents challenges including motion blur, appearance change, complex deformation, occlusion, slow motion, and interacting objects.
\textbf{FBMS-59}~\cite{ochs2013segmentation} has 59 sequences with greatly varied resolution and annotates every 20th frame. Many sequences contain multiple moving objects. Following previous evaluation metric~\cite{Yang_2019_CVPR, xie2022segmenting}, we merge objects of SegTrackv2 and FBMS-59 into one single object for video object segmentation.
We evaluate the pixel-wise segmentation through Jaccard index $\mathcal{J}$, also called Intersection over Union (IoU). Following prior arts~\cite{Yang_2019_CVPR, Yang_2021_ICCV}, we compute the mean per frame over the test set and merge multi-object annotation into single unified segmentation.
% We do not apply any post-processing trick e.g., CRF, temporal smoothing, and spectral clustering. 
% \weidi{is this still true ?}

\subsection{Implementation Details}
\label{imp}
For data input, we sample $T=7$ consecutive frames as the input clip. 
% \weidi{I don't understand the definition of stride, do you mean temporal stride ? 
% why not simply say what is the fps?} 
% \dsr{It means sampling every other frames. Maybe directly erasing would be better.}
Each frame is resized to $192 \times 384$ and the estimated optical flow is computed by RAFT~\cite{teed2020raft}, which is pre-trained on the synthetic dataset~\cite{MIFDB16}. 

To compute spatial-temporal visual representation, 
we adopt the first three stages of a SwinV2-T as the frame encoder $\Phi_{\text{enc}}$, 
which is then followed by a standard Transformer Encoders with 8 heads~\cite{vaswani2017attention} as temporal fusion module $\Phi_{\text{temp}}$.
% \weidi{just one layer ? with how many heads ?} \dsr{The gain compared 3 layers to 1 is not significant in DAVIS but much slower and more paras.}
For the frame comparator $\Phi_{\text{comp}}$, we use two deformable convolutional layers in the company with three standard Transformer Encoders with 8 heads to process the pixel transformation. 
Then, we choose $N=5$ iteration in total for the iterative routing in slot attention. 
Lastly, we utilize three stages of SwinV2 blocks with the linear patch expanding layers as the flow decoder $\Phi_{\text{dec}}$~\cite{cao2021swin}.

As for training, we adopt AdamW optimizer~\cite{loshchilov2018decoupled} with learning rate $4\times 10^{-5}$. 
The model is trained from scratch without any pretrained weights, for a total of 300k iterations. At inference time, we adopt the temporal sliding window to ensemble the segmentation masks. We average the resultant masks obtained by the whole temporal segments. We propose two protocols to evaluate our results. Besides measuring the masks without any post-processing, we also apply \textbf{test-time adaptation} with the help of the self-supervised DINO-pretrained ViT~\cite{caron2021emerging}. Without any fine-tuning, the pretrained ViT can propagate the masks as noisy annotations to the whole frames in the same manner as CRW~\cite{jabri2020space}. We refine the masks further with CRF~\cite{lafferty2001conditional}. For more detailed technical information, please refer to the supplementary materials. 
% To show the pure quality of our generated masks, we do not apply any post-processing trick e.g., CRF, temporal smoothing, and spectral clustering. 

\section{Results}
In this section, 
we compare primarily with several top-performing approaches trained without human annotations, for example, OCLR~\cite{xie2022segmenting}, MG~\cite{Yang_2021_ICCV}, SIMO~\cite{lamdouar2021segmenting}, CIS~\cite{Yang_2019_CVPR}, etc.
However, as the architecture, modality, input resolution, and post-processing protocols are all different, we try our best to conduct the comparison as fairly as possible. 

% we first dissect the key components of our methods as ablation study~\ref{abl}.
\subsection{Ablation Study}
\label{abl}
We conduct all ablation studies on DAVIS2016 and vary one variable each time, as shown in Table~\ref{tab:abl}.

\begin{table}[!htb]
    \centering
    \small
    \begin{tabular}{cccccccc}
    \toprule
         Model & $\Phi_{\text{temp}}$ & $\Phi_{\text{comp}}$ & T & $\mathcal{L_{\text{cons}}}$  & $\mathcal{L_{\text{entro}}}$   & DAVIS($\mathcal{J}$ $\uparrow$) \\\midrule
         Ours-A & \checkmark & \checkmark  & 7  & \checkmark & \checkmark & \textbf{73.9} \\\midrule
         Ours-B &  - & \xmark  & 7  &\checkmark & \checkmark  &  fail \\
         Ours-C & \xmark & \checkmark & 7 & \checkmark & \checkmark   &   68.3 \\\midrule
         Ours-D &  \checkmark & \checkmark & 3  & \checkmark & \checkmark   & 66.4 \\
         Ours-E &  \checkmark & \checkmark & 5  & \checkmark & \checkmark  & 68.2 \\\midrule
         Ours-F & \checkmark & \checkmark & 7 & \xmark & \xmark  &  60.4 \\
         Ours-G &  \checkmark & \checkmark & 7 & \xmark & \checkmark & 65.6 \\
         Ours-H & \checkmark & \checkmark & 7 & \checkmark & \xmark &  69.5 \\

    \bottomrule     
    \end{tabular}
    \caption{Ablation studies on temporal fusion ($\Phi_{\text{temp}}$), frame comparator ($\Phi_{\text{comp}}$), the number of input frames ($T$), temporal consistency ($\mathcal{L_{\text{cons}}}$), and entropy loss ($\mathcal{L_{\text{entro}}}$).}
    % \weidi{re-order the column by importance, entropy is clearly not the most critical one.}
    \label{tab:abl}
\end{table}

\paragraph{Temporal Fusion $\Phi_{\text{temp}}$ and Frame Comparator $\Phi_{\text{comp}}$.}
As shown by Ours-A and Our-C, the performance degrades significantly without temporal fusion, demonstrating the importance of building up global temporal dependency. Also, indicated by Ours-B, we find the model fails to converge when removing the component of frame comparator $\Phi_{\text{comp}}$. It meets expectations because frame comparator $\Phi_{\text{comp}}$ is the sole module in charge of relative motion estimation. Without it, the model cannot reconstruct optical flow thus leading to divergence. 

\paragraph{Number of Frames $T$.}
While comparing Ours-A, Ours-D, and Ours-E, 
there is a clear trend that increasing the frame number boosts the segmentation quality, which coincides with our intuition that incorporating a wider temporal receptive field can enhance the sense of temporal coherence and object permanence. Due to the limited computational memory, 
we only set $T=7$ as the maximum frame number in the paper.
A promising performance is expected when inputting more frames.

\paragraph{Temporal Consistency $\mathcal{L_{\text{cons}}}$ and Entropy Regularisation $\mathcal{L_{\text{entro}}}$.}
Lastly, comparing Ours-G and Ours-A, we observe that the performance increases considerably with temporal consistency. It manifests the validity of our design motivation, temporal consistency conduces to persistently tracking the object. The entropy regularisation is also indispensable shown by Ours-H and Ours-A.

\subsection{Comparison with State-of-the-art}
\label{sota}
We show the comparison with state-of-the-art in Table~\ref{tab:sota}. On DAVIS2016, MOD achieves 73.9\% mIOU without any post-processing, exceeding MG~\cite{Yang_2021_ICCV} by a large margin (+5.8\%). Compared to the latest method OCLR~\cite{xie2022segmenting} which fabricates a synthesized dataset to train its model, our method still surpasses it only using the information of the DAVIS dataset itself. On another two benchmarks SegTrackv2 and FBMS-59, MOD also beats MG~\cite{Yang_2021_ICCV}, which only leverages single-step flows to decompose foreground and background, by +3.6\% and +8.2\%, respectively. The superior experimental results demonstrate that our methodology of multi-frame reasoning benefits moving object discovery.  
Furthermore, equipped with DINO-pretrained ViT, a further performance gain is observed on all three benchmarks, which is even more competitive with current supervised approaches. 

% \begin{table}[!htb]
%     \centering
%     \begin{tabular}{lll}
%     \toprule
%          Method & static rep. & ensemble \\\midrule
%          w/o $\mathcal{L_{\text{ind}}}$ & 59.7 & 70.5 \\
%          w/ $\mathcal{L_{\text{ind}}}$ & 67.7 ($8.0 \uparrow$)&  73.7 ($3.2 \uparrow$) \\
%     \bottomrule     
%     \end{tabular}
%     \caption{Ablation studies on motion inductive loss $\mathcal{L_{\text{ind}}}$. We report the average of region similarity ($\mathcal{J}$) over the DAVIS 2016 evaluation set. \textit{Static rep.} means the predicted masks generated by replicating frames while \textit{ensemble} means we evaluate the average of multiple masks including dynamic flow and static replication.
%     \weidi{this table is useless, right ?}}
%     \label{tab:static}
% \end{table}
\label{exp}

\begin{table*}[t]
\setlength\tabcolsep{6pt}
  \label{single1-table}
  \centering
  \begin{tabular}{cccccccc}  
    \toprule
    {} & \multicolumn{2}{c}{Training} & \multicolumn{2}{c}{Inference} & \multicolumn{3}{c}{$\mathcal{J}$ (Mean)  $\uparrow$ }  \\
    \cmidrule(r){2-3}
    \cmidrule(r){4-5}
    \cmidrule(r){6-8}
    Model   & HA & Sup. & RGB &  Flow  & DAVIS2016 & SegTrackv2 & FBMS-59 \\ \midrule
    SAGE~\cite{wang2017saliency} & \xmark &  None &  \checkmark &  \checkmark & 42.6 & 57.6 & 61.2 \\
    NLC~\cite{faktor2014video} & \xmark &  None & \checkmark &  \checkmark & 55.1  & 67.2  & 51.5 \\
    CIS~\cite{Yang_2019_CVPR} & \xmark &  None & \checkmark &  \checkmark & 71.5 & 62.5 & 63.5 \\
    AMD~\cite{liu2021emergence}  & \xmark &  None &  \checkmark & \xmark & 57.8 & 57.0 & 47.5 \\
    SIMO~\cite{lamdouar2021segmenting} & \xmark &  None & \xmark &  \checkmark & 67.8  & 62.0  & - \\
    MG~\cite{Yang_2021_ICCV}  & \xmark &  None & \xmark & \checkmark &  68.3 & 58.6 & 53.1 \\
    OCLR~\cite{xie2022segmenting} & \xmark  & Syn. & \xmark & \checkmark &  72.1 & 67.6 & 65.4 \\ 
    \textbf{MOD (w/o post-processing)} & \xmark & None & \checkmark & \xmark & 73.9 & 62.2 & 61.3 \\
    \textbf{MOD (test-time adaptation)} & \xmark & None & \checkmark & \xmark & 79.2 & 69.4 & 66.9 \\ \midrule
    FSEG~\cite{dutt2017fusionseg} & \checkmark & GT & \checkmark & \checkmark  &  70.7 & 61.4 & 68.4 \\
    COSNet~\cite{lu2019see} & \checkmark & GT & \checkmark & \xmark &  80.5 & 49.7 & 75.6  \\
    MATNet~\cite{zhou2020motion} & \checkmark & GT & \checkmark & \checkmark &  82.4 & 50.4 & 76.1 \\
    % 3DC-Seg~\cite{Mahadevan20BMVC} & \checkmark & GT & \checkmark & \xmark &  84.3 & - & - \\
    D$^2$Conv3d~\cite{schmidt2022d2conv3d} & \checkmark & GT & \checkmark & \xmark &  85.5 & - & - \\
    \bottomrule
  \end{tabular}
    \caption{Quantitative comparison on unsupervised video object segmentation. We compare our method on three standard datasets, DAVIS2016, SegTrackv2, and FBMS-59. HA is short for \textbf{Human Annotation}. Sup. refers to the supervision, including None, Synthetic (Syn.), and Ground Truth (GT).}
  \label{tab:sota}
\end{table*}

\subsection{Qualitative Results}
\begin{figure*}[!htb]
    \centering
    \includegraphics[width=\textwidth]{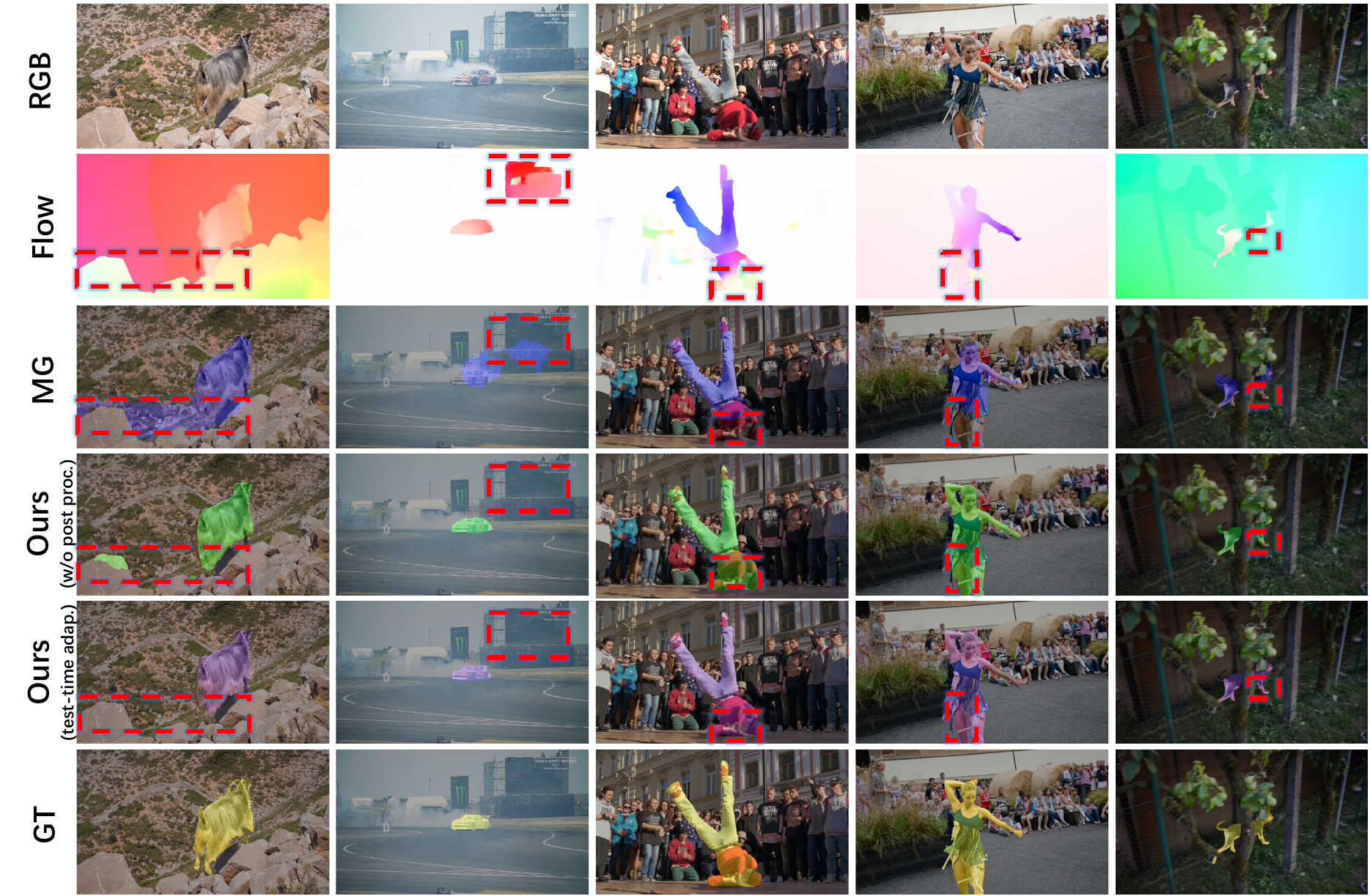}
    \caption{Qualitative results of object video segmentation on DAVIS2016. MG refers to \cite{Yang_2021_ICCV}. Red boxes outline the corresponding difference.}
    % \weidi{is it possible to do another figure, with sequences where objects stop moving at intermediate step, and your method still works.}
    \label{fig:blend}
\end{figure*}

In Figure~\ref{fig:blend}, we present several qualitative illustrations of the model. It can be seen that our results are robust to the noticeable background flow signal (drift-chicane sequence in the second column) and estimate more accurate boundaries when single-step foreground flow cannot represent exact object shape (breakdance and dance-twirl sequences in middle) compared to MG~\cite{Yang_2021_ICCV}. It demonstrates inferring the masks by associating a bunch of RGB features well resolves the limitation of the usage of flow. Moreover, in virtue of temporally consistent cues, our model handles occlusion well shown in the libby sequence at the rightmost column, which could be hard for the flow-only method to maintain the object shape constantly.

\subsection{Limitations}
Though we demonstrate that associating multiple frames stimulates the comprehensive sense of objectness, which is proved by the superior experimental performance across the prior arts, there still exist limitations and room for improvement. \textit{First}, our method uses a number of consecutive frames, which challenges computational memory. How to utilize pretrained features for reducing training expenses would be meaningful.  \textit{Second}, how to segment multiple objects remains unexplored. It will be promising to exploit the semantic information from RGB to discriminate different foreground objects. We leave them as the feature work.
Despite these limitations, the approach has convincingly manifested the value of considering textural information and processing RGB frames as a whole.

\section{Related Work}
\label{rw}
\paragraph{Video Object Segmentation.} How to segment objects coherently in one video sequence has extended the topic of instance segmentation in the image. There is a great amount of work about video object segmentation (VOS) in recent decades~\cite{caelles2017one, hu2017maskrnn, fan2019shifting, dutt2017fusionseg, Lai19, maninis2018video, oh2019video, voigtlaender2019feelvos, caelles2017one, perazzi2017learning, hu2018videomatch, li2018video, bao2018cnn, voigtlaender2019feelvos, johnander2019generative}. Recently, the research on getting rid of the dense annotation and designing more effective self-supervised algorithms has attracted more and more interest in the computer vision community including VOS~\cite{xu2021rethinking, jabri2020space, lai2020mast, NEURIPS2019_140f6969, vondrick2018tracking, Lu_2020_CVPR, wang2019unsupervised, kipf2022conditional}.
For VOS, there are two mainstream protocols to evaluate the learned model. One is semi-supervised video object segmentation, the other is unsupervised video object segmentation. Given the first-frame mask of the objects of interest, semi-supervised VOS tracks those objects in subsequent frames, while unsupervised VOS directly segments the most salient objects from the background without any reference. These two protocols are defined in the inference phase, meaning methods could leverage ground truth annotations in the training stage. In this paper, we don't use any kinds of manual annotations for either training or evaluation.

\paragraph{Motion Segmentation.} As the name suggests, the aim of motion segmentation is to discover moving objects. 
One line of the work~\cite{brox2010object, fragkiadaki2012video, ochs2012higher, ochs2013segmentation, lezama2011track, keuper2015motion} formulates motion as the point trajectory to take advantage of long-range temporal information so that segmentation results can be acquired by grouping the trajectories. 
Later, deep learning methods take over the area~ \cite{tokmakov2017learninga, tokmakov2017learning, xie2019object, Yang_2019_CVPR, Yang_2021_ICCV, choudhury2022guess}. \cite{tokmakov2017learninga} adopts a two-stream network that ingests both RGB and optical flow. Then they realize a memory mechanism by the convolutional recurrent unit to enhance the visual cues. CIS~\cite{Yang_2019_CVPR} achieves fully unsupervised motion segmentation which discards the supervision of annotated masks during training. By formulating a min-max game of mutual information, the generator is asked to create foreground segments that are as unrelated as possible to the background. AMD~\cite{liu2021emergence} minimizes the warp synthesis error to train appearance and motion pathways without any supervision.
MG~\cite{Yang_2021_ICCV} solely leverages the optical flow to separate the pixels via cross attention mechanism~\cite{locatello2020object}. Recently, OCLR~\cite{xie2022segmenting} synthesizes a training dataset with multiple moving objects systematically to train their model. Though our method is a bit like MG, the key difference in our work is that we purely take RGB frames as input thus relieving the over-reliance on optical flow and allowing us to segment objects at test time when optical flow is not accessible. \\
\paragraph{Object Discovery.} There is rich literature on identifying salient objects without explicit supervision, known as object discovery.
There exist a series of works that aim to learn object-centric representations in images~\cite{locatello2020object,lin2020space,jiang2020generative,greff2019multi,emami2021efficient,burgess2019monet,crawford2019spatially,engelcke2019genesis,engelcke2021genesis}. Typically, IODINE~\cite{greff2019multi} develops iterative variational inference to separate different objects. \cite{locatello2020object} proposes slot attention to iteratively update latent object representations. Further, a line of works~\cite{zablotskaia2020unsupervised,min2021gatsbi,kosiorek2018sequential,kipf2022conditional,jiang2019scalor,kabra2021simone,crawford2020exploiting,besbinar2021self,bear2020learning,Bao_2022_CVPR, ye2022sprites, Yang_2021_ICCV} extend object-centric learning to video domain. Most of these approaches incorporate motion cues into the reconstruction task and perform well in moving object segmentation but perform poorly in processing static objects. While in this work, we adopt slot attention to form optical flow reconstruction bottleneck and perceive both dynamic and static instances through RGB clip input in a temporally consistent way.

\section{Conclusion}
% \cite{kipf2022conditional,Yang_2019_CVPR,liu2021emergence}
In this paper, we propose a self-supervised model for video object discovery.
The model takes a set of consecutive RGB frames as input and generates the segmentation mask for the moving objects in the video. 
At training time, the model is tasked to reconstruct the optical flow between any pair of frames, through a layered representation with the opacity channels being treated as the segmentation.
To encourage the model to capture the objects even when they may be static at a certain time point, a temporal consistency loss is enforced on the inferred masks on the randomly-paired frames. As a consequence, we demonstrate superior performance over previous state-of-the-art methods on three public video segmentation datasets (DAVIS2016, SegTrackv2, and FBMS-59), while being computationally efficient by avoiding the overhead of computing optical flow.


{\small
\bibliographystyle{ieee_fullname}
\bibliography{egbib}
}

\appendix
\section{More Implementation Details}
In this section, we list the architecture details and training settings. 
Codes and models will be released publicly.
\subsection{Visual Encoder}
The visual encoder contains the first three stages of Swin-Tiny V2. We tabulate the workflow in Table~\ref{tab:enc}. 
\begin{table}[h]
    \centering
    \begin{tabular}{c c c}
    \toprule
         stage & operation & output sizes\\\hline
         input & - & 3 $\times$ 192 $\times$ 384 \\
         PatchEmbed & 4 $\times$ 4, stride 4, 96 & 96 $\times$ 48 $\times$ 96 \\
         SwinBlock1 &  $ \begin{bmatrix} 
                        h=3\\
                        ws=12
                        \end{bmatrix}
                    \times 2 $ & 96 $\times$ 48 $\times$ 96 \\
         DownSample1 &  192 & 192 $\times$ 24 $\times$ 48 \\
         SwinBlock2 &   $ \begin{bmatrix} 
                        h=6\\
                        ws=12
                        \end{bmatrix}
                    \times 2 $ & 192 $\times$ 24 $\times$ 48 \\
         DownSample2 &  384 & 384 $\times$ 12 $\times$ 24 \\
         SwinBlock3 &   $ \begin{bmatrix} 
                        h=12\\
                        ws=12
                        \end{bmatrix}
                    \times 6 $ & 384 $\times$ 12 $\times$ 24 \\
    \bottomrule
    \end{tabular}
    \caption{Architecture of visual encoder. $h$ stands for the number of attention heads while $ws$ refers to window size.}
    \label{tab:enc}
\end{table}
\subsection{Frame Comparator}
The frame comparator possesses two deformable convolutional layers, with ReLU operation in between and three Transformer Encoder blocks. We tabulate the workflow in Table~\ref{tab:deform}. 
\begin{table}[h]
    \centering
    \begin{tabular}{c c c}
    \toprule
         stage & operation & output sizes\\\hline
         input & - & 768 $\times$ 12 $\times$ 24 \\
         DeformConv1  & 3 $\times 3$, 768 & 768 $\times$ 12 $\times$ 24 \\
         ReLU & - & 768 $\times$ 12 $\times$ 24 \\
         DeformConv2  & 3 $\times 3$, 384 & 384 $\times$ 12 $\times$ 24 \\
         Transformer  & $\begin{bmatrix} 
                        h=8\\
                        384
                        \end{bmatrix} \times 3$ & 384 $\times$ 12 $\times$ 24 \\
    \bottomrule
    \end{tabular}
    \caption{Architecture of frame comparator. $h$ represents the number of attention heads}
    \label{tab:deform}
\end{table}

\subsection{Flow Decoder}
The frame comparator consists of three stages of SwinV2 block added with linear expanding layers. We tabulate the workflow in Table~\ref{tab:dec}. 

\begin{table}[h]
    \centering
    \begin{tabular}{c c c}
    \toprule
         stage & operation & output sizes\\\hline
         input & - & 384 $\times$ 12 $\times$ 24 \\
          SwinBlock1 &   $ \begin{bmatrix} 
                h=12\\
                ws=12
                \end{bmatrix}
            \times 2 $ & 384 $\times$ 12 $\times$ 24 \\
           PatchExpand1 & 768 & 192 $\times$ 24 $\times$ 48 \\
          SwinBlock2 &   $ \begin{bmatrix} 
                h=6\\
                ws=12
                \end{bmatrix}
            \times 2 $ & 192 $\times$ 24 $\times$ 48 \\
           PatchExpand2 & 384 & 96 $\times$ 48 $\times$ 96 \\
          SwinBlock3 &   $ \begin{bmatrix} 
                h=3\\
                ws=12
                \end{bmatrix}
            \times 2 $ & 96 $\times$ 48 $\times$ 96 \\
        PatchExpand3 & 1536 & 96 $\times$ 192 $\times$ 384 \\
        outConv & $5 \times 5$, stride 1, 4 & 4 $\times$ 192 $\times$ 384 \\
    \bottomrule
    \end{tabular}
    \caption{Architecture of flow decoder. $h$ stands for the number of attention heads while $ws$ refers to window size.}
    \label{tab:dec}
\end{table}
\subsection{Training Details}
For all datasets, we train with a batch size of 2. To train more efficiently, we sample three flow pairs ($i \rightarrow j$) for a given frame $i$; one is static replication $i=j$, another two are motion pair $i \neq j$. We apply temporal consistency first on the masks from two motion pairs, then pull the static mask to the average of two dynamic masks closer. We linearly warm up the learning rate for the first 1k iterations. Besides, for every 100k iterations, we decay the learning rate by half and increase the scale of temporal consistency $\lambda_{c}$ and entropy regularisation $\lambda_{e}$ by the factor of $5$. In the default setting, we train for about 3 days on 8 standard Tesla V100 GPUs with 32GB memory each.

\subsection{Test-time Adaptation}
Inspired by OCLR~\cite{xie2022segmenting}, we adopt test-time adaptation based on RGB sequence to enhance appearance consistency. In detail, we follow existing works on self-supervised tracking~\cite{lai2020mast,jabri2020space} to propagate object masks across time span. The whole process consists of three steps. First, we extract RGB features of each frame with a DINO-pretrained ViT encoder. Then, we select key frames for object mask propagation. Finally, we calculate the affinity matrix between frames and perform mask propagation.

\textbf{DINO Feature Extraction.} Given a video sequence $v=\{x_1,...,x_T\}, v\in\mathbb{R}^{T\times H\times W\times 3}$, we use DINO pretrained ViT-Small encoder with patch size $8\times 8$ to extract features:
\begin{align}
    \{f_1,...,f_T\} = \{\Phi(x_1),...,\Phi(x_T)\},
\end{align}
where $f_t\in\mathbb{R}^{h\times w\times 384}$, $h=H//8$ and $w=W//8$. The extracted features will be used in the mask propagation step.

\textbf{Key Frame Selection.} Given video $v$, our model predicts object mask of each frame as $m=\{\alpha_1,...,\alpha_T\}, m\in\mathbb{R}^{T\times H\times W\times 1}$. Since the video frames are continuous along the temporal dimension, it is practical to propagate object masks between neighboring frames. The propagation operation is the same as~\cite{jabri2020space}, the only difference is that we have no ground-truth mask for reference. Therefore, we need to design a mechanism to select object masks of high confidence. To do this, we measure the temporal coherence of predicted object masks for key frame selection. Specifically, for each timestamp $t\in\{3,...,T-2\}$, we can calculate four propagated masks as:
\begin{align}
    \hat{\bm{\alpha}}_t=[\text{Mask-prop}(\alpha_{t-i})]_{i\in \{-2, -1, 1, 2\}},
\end{align}
where `Mask-prop' denotes the propagation operation.
Then we calculate the average IoU between the original mask and propagated masks as the confidence score:
\begin{align}
    s_t = \frac{1}{4} \sum_{i=0}^3 \text{IoU}(\hat{\bm{\alpha}}_t[i],\alpha_t).
\end{align}
The calculated $s_t$ measures the coherency between $\alpha_t$ and its neighbors. Our empirical studies show that it serves as a reliable signal for key frame selection.

\textbf{Object Mask Propagation.}
We select timestamps with Top-$k\%$ confidence score as the key reference frames ($k=15$ on DAVIS2016, $k=25$ on SegTrackv2, $k=10$ on FBMS-59). Then we iteratively propagate the object masks with a neighbor temporal window size $n=7$.
\section{Results Breakdown}
We include a specific result breakdown in this section. We show the per-sequence results on DAVIS2016 in Table~\ref{tab:davis}, SegTrackv2 in Table~\ref{tab:stv2} and FBMS-59 in Table~\ref{tab:fbms}.
\begin{table}[!htb]
\centering
\begin{tabular}{ccc}
\toprule
        & \multicolumn{2}{c}{$\mathcal{J}$ (Mean)  $\uparrow$}\\
        \cmidrule(r){2-3}
        Sequence &  w/o post proc. & test-time adap. \\\hline
        dog &  80.7 &  87.4 \\
        cows & 87.2 & 88.8 \\
        goat & 47.5 & 80.6 \\
        camel & 85.6 & 86.1 \\
        libby & 72.5 & 77.7\\
        parkour & 72.9 & 87.9\\
        soapbox & 84.6  & 86.5\\
        blackswan & 48.9 &  46.9\\
        bmx-trees & 50.1 & 55.8 \\
        kite-surf &  55.9 & 62.6 \\
        car-shadow & 87.9  &  86.9\\
        breakdance & 82.6  & 76.0\\
        dance-twirl & 82.5 & 85.4 \\
        scooter-black & 80.2 & 80.3 \\
        drift-chicane & 78.6 & 82.2\\
        motocross-jump & 68.4 & 88.9 \\
        horsejump-high & 78.0 & 84.3 \\
        drift-straight & 69.2 & 80.0 \\
        car-roundabout & 87.7 & 83.9 \\
        paragliding-launch & 62.1 & 62.8 \\\hline
        % seq. avg. & 68.7 \\
        frame avg. & 73.9 & 79.2 \\\bottomrule 
\end{tabular}
\caption{Sequence-wise results on DAVIS2016.}
\label{tab:davis}
\end{table}

\begin{table}[!htb]
\centering
\begin{tabular}{ccc}
\toprule
        & \multicolumn{2}{c}{$\mathcal{J}$ (Mean)  $\uparrow$}\\
        \cmidrule(r){2-3}
        Sequence &  w/o post proc. & test-time adap. \\\hline
        drift & 41.7 & 40.7 \\
        birdfall & 38.2 & 61.5 \\
        girl &  76.5 & 82.3 \\
        cheetah &  18.4 & 30.1 \\
        worm & 52.5 & 74.3 \\
        parachute & 90.2 & 92.0 \\
        monkeydog & 14.3 & 31.1\\
        hummingbird & 61.2 & 58.8\\
        soldier & 66.3 & 58.8 \\
        bmx &  73.7 & 78.8 \\
        frog & 80.5 & 76.3 \\
        penguin & 63.5 & 62.7 \\
        monkey & 46.8 & 77.4\\
        bird of paradise & 85.3 & 85.4
        \\\hline
        frame avg. & 62.2  & 69.4 \\\bottomrule 
\end{tabular}
\caption{Sequence-wise results on SegTrackv2.}
\label{tab:stv2}
\end{table}

\begin{table}[!htb]
\centering
\begin{tabular}{ccc}
\toprule
        & \multicolumn{2}{c}{$\mathcal{J}$ (Mean)  $\uparrow$}\\
        \cmidrule(r){2-3}
        Sequence &  w/o post proc. & test-time adap. \\\hline
        camel01 & 25.8 & 66.9 \\
        cars1 & 66.1  & 88.3\\
        cars10 & 31.6 & 33.9\\
        cars4 &  72.9 & 83.2  \\
        cars5 &  81.2 & 82.5 \\
        cats01 &  80.6  &  79.2 \\
        cats03 & 62.0  &  63.4  \\
        cats06 & 40.1 & 38.7 \\
        dogs01 & 70.6 & 61.2  \\
        dogs02 &  62.8 &  82.2 \\
        farm01 & 86.7  & 88.9 \\
        giraffes01 & 38.6 & 52.2  \\
        goats01 & 44.8  & 45.3 \\
        horses02 & 64.4  & 77.6 \\
        horses04 & 68.5  & 73.5 \\
        horses05 & 43.7 & 49.0  \\
        lion01 & 60.1 & 71.5 \\
        marple12 & 74.7 & 80.3 \\
        marple2 & 65.0 & 71.4 \\
        marple4 & 79.1 & 91.6\\
        marple6 & 76.0  & 85.1 \\
        marple7 & 72.5 & 55.2 \\
        marple9 & 87.5 & 97.9 \\
        people03 & 76.5 & 48.5\\
        people1 & 72.4  &  80.8 \\
        people2 &  80.9  &  83.0 \\
        rabbits02 & 50.2 & 58.9 \\
        rabbits03 & 39.5  & 55.7 \\
        rabbits04 & 47.0 & 53.0 \\
        tennis & 63.1 & 71.1 \\\hline
        frame avg. & 61.3 & 66.9 \\\bottomrule 
\end{tabular}
\caption{Sequence-wise results on FBMS-59.}
\label{tab:fbms}
\end{table}

\section{More Qualitative Results}
\begin{figure*}
    \centering
    \includegraphics[width=\textwidth]{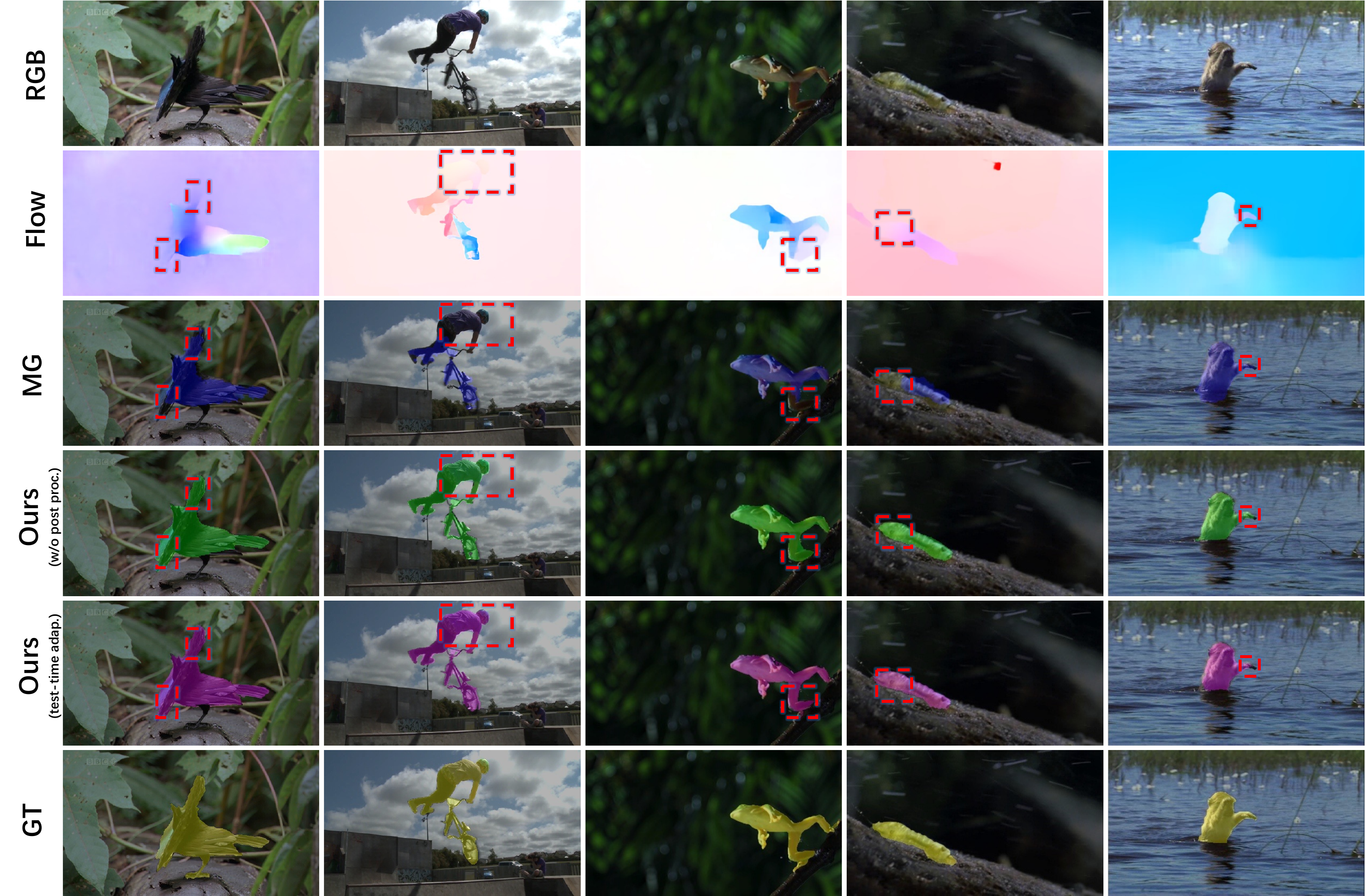}
    \caption{Qualitative results on SegTrackv2. MG refers to \cite{Yang_2021_ICCV}. Red boxes outline the corresponding difference.}
    \label{fig:stv2}
\end{figure*}

\begin{figure*}
    \centering
    \includegraphics[width=\textwidth]{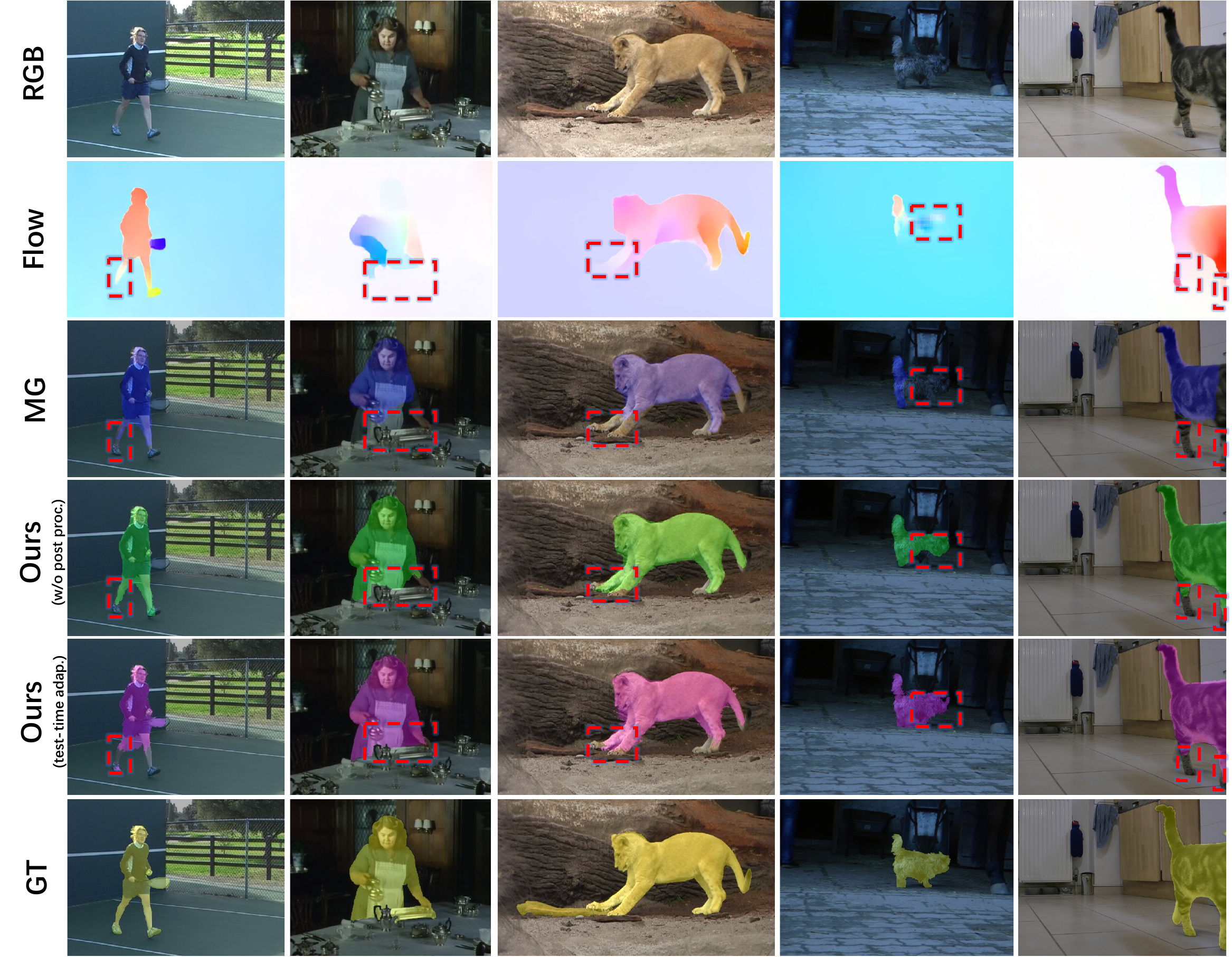}
    \caption{Qualitative results on FBMS-59. MG refers to \cite{Yang_2021_ICCV}. Red boxes highlight the corresponding difference.}
    \label{fig:fbms}
\end{figure*}

We show more qualitative results of SegTrackv2 and FBMS-59 in Figure~\ref{fig:stv2} and Figure~\ref{fig:fbms}.
\end{document}

\title{Supplementary Material for Motion-aware Contrastive Video Representation Learning via Foreground-background Merging}
\author {
    % Authors
    Shuangrui Ding\textsuperscript{\rm 1}\thanks{Work done during an internship at Tencent AI Lab.} \quad
    Maomao Li\textsuperscript{\rm 2} \quad Tianyu Yang\textsuperscript{\rm 2}  \quad Rui Qian\textsuperscript{\rm 3} \\Haohang Xu\textsuperscript{\rm 1}  \quad Qingyi Chen\textsuperscript{\rm 4} \quad Jue Wang\textsuperscript{\rm 2} \quad 
    Hongkai Xiong\textsuperscript{\rm 1}\thanks{ Corresponding author. Email: xionghongkai@sjtu.edu.cn.}\\
    \textsuperscript{\rm 1}Shanghai Jiao Tong University \quad 
    \textsuperscript{\rm 2}Tencent AI Lab  \\  \textsuperscript{\rm 3}The Chinese University of Hong Kong  \textsuperscript{\rm 4}University of Michigan \\
    {\tt\small \{dsr1212, xuhaohang, xionghongkai\}@sjtu.edu.cn tianyu-yang@outlook.com}\\
    {\tt\small \{limaomao07, arphid\}@gmail.com  qr021@ie.cuhk.edu.hk chenqy@umich.edu}
}
\maketitle
\section{More Implementation Details}
\subsection{Self-supervised Pretraining Details.}

In the pretraining stage, we adopt the SGD optimizer with the initial learning rate of 0.01 and weight decay of $10^{-4}$, and we decay the learning rate by 0.1 at epoch 120 and 180. For the implementation of MoCo, we closely follow the parameter setting in~\cite{chen2020improved}. The number of the negative queue is set to 65536 for Kinetics-400, and 2048 for UCF101, respectively. We also swap the key/queue samples so that each sample can generate the gradient for optimization. The momentum of updating the key encoder is 0.999, and the temperature hyper-parameter $\tau$ is 0.1. We use a 2-layer MLP projection head.
\subsection{Augmentation Details.}
We perform data augmentation using Kornia package~\cite{riba2020kornia}.
In the pretraining and finetune phase, we crop 224$\times$224 or 112$\times$112 pixels from a video with RandomResizedCrop, which randomly resizes the input area between a lower bound and upper bound. We set the bound as [0.2, 1]. Then, the basic augmentation set consists of RandomGrayscale (probability 0.2), ColorJitter (probability 0.8, \{brightness, contrast, saturation, hue\} = \{0.4, 0.4, 0.4, 0.1\}), RandomHorizontalFlip (probability 0.5) and RandomGaussianBlur (probability 0.5, the kernel with radius 23 and standard deviation $\in$ [0.1, 2.0]). In the linear probe stage, we take a simpler augmentation setting instead. We only apply RandomResizedCrop with the bound [0.2, 1] and RandomHorizontalFlip (probability 0.5).

\subsection{More Details on Action Recognition.}
 In the finetune stage, the SGD optimizer is adopted with the initial learning rate of 0.025 and weight decay of $10^{-4}$. We finetune the model for 150 epochs with a batch size of 128 on 4 Tesla V100 GPUs. We decay the learning rate by 0.1 at epoch 60 and 120. Besides, we add the dropout layer before the last fully connected layer. We set dropout rate 0.7 for UCF101 and 0.5 for HMDB51, respectively. 
 
We train the last fully connected layer in the linear probe with the initial learning rate of 5 and weight decay of 0. We finetune the model for 100 epochs with a batch size of 128 on 4 Tesla V100 GPUs. We decay the learning rate by 0.1 at epoch 60 and 80. Besides, We $\mathcal{L}_2$ normalize the embeddings before the last fully connected layer. 
\section{More Visualization of FAME}
\begin{figure}
    \centering
    \includegraphics[width=\linewidth]{Figure/FAME.png}
    \caption{Illustration of FAME visualization. The first row is the video frame while the second row is the foreground mask FAME generates.}
    \label{fig:FAME}
\end{figure}
In Figure~\ref{fig:FAME}, we show more foreground masks obtained from FAME. We show that FAME can discover most regions of the foreground objects and remove the monotonous backgrounds. 

\section{CAM Visualization}

Besides CAAM visualization, we provide the CAM~\cite{zhou2016learning} visualization in Figure~\ref{fig:cam}.
% We train a linear classifier with the pretrained model similar to the linear probe and omit the last global average pooling layer to generate activation maps. 
With that, we can spot the contribution of each area and find crucial regions for discriminating the specific action class. We find that when integrated with FAME, the model can focus on moving foreground area rather than background context. For example, in the first row of Figure~\ref{fig:cam}, FAME precisely captures the moving upper and lower body when the man is practicing TaiChi, while the baseline displays a dispersed highlight map and fails to attend to the motion area. In addition, we illustrate that the CAM activation map can almost overlap with the foreground mask generated by FAME. It testifies that our strong motion inductive augmentation guides the model to perceive the motion patterns and hinder the background bias. 

\begin{figure}
    \centering
    \includegraphics[width=\linewidth]{Figure/cam.png}
    \caption{Class activation maps (CAM) visualization. Red areas indicate the important areas for the model to predict the action class. Comparing to baseline, FAME resists the impact of background and highlights the motion areas.}
    \label{fig:cam}
\end{figure}
\section{Visualization of Video Retrieval}
In Figure~\ref{fig:RET}, we demonstrate the results of video retrieval. After pretraining the model on Kinetics-400, we conduct the video retrieval experiment on UCF101. The results show that our model can retrieve diverse video samples that share the same action semantics with the query, regardless of the background context. For example, in Fig~\ref{fig:longjump}, the query sample contains the action in the sandpit, and our model could retrieve the long jump samples in the standard stadium. Though the backgrounds in the query and retrieved videos are quite different, our model achieves accurate retrieval by attending to the dynamic motions and understanding the true action semantics.
\begin{figure}
    \centering
    \subfloat[BandMarching]{\includegraphics[width=0.25\linewidth]{retrieve/90/q.jpg}
    \hspace{2mm}
    \includegraphics[width=0.25\linewidth]{retrieve/90/k_0.jpg}
    \includegraphics[width=0.25\linewidth]{retrieve/90/k_1.jpg}
    \includegraphics[width=0.25\linewidth]{retrieve/90/k_2.jpg}}
    \hspace{4mm}\\
    \subfloat[BenchPress]{\includegraphics[width=0.25\linewidth]{retrieve/198/q.jpg}
    \hspace{2mm}
    \includegraphics[width=0.25\linewidth]{retrieve/198/k_0.jpg}
    \includegraphics[width=0.25\linewidth]{retrieve/198/k_1.jpg}
    \includegraphics[width=0.25\linewidth]{retrieve/198/k_2.jpg}}
    \hspace{4mm}\\
    \subfloat[HorseRace]{\includegraphics[width=0.25\linewidth]{retrieve/784/q.jpg}
    \hspace{2mm}
    \includegraphics[width=0.25\linewidth]{retrieve/784/k_0.jpg}
    \includegraphics[width=0.25\linewidth]{retrieve/784/k_1.jpg}
    \includegraphics[width=0.25\linewidth]{retrieve/784/k_2.jpg}}
    \hspace{4mm}\\
    \subfloat[LongJump]{\includegraphics[width=0.25\linewidth]{retrieve/998/q.jpg}
    \hspace{2mm}
    \includegraphics[width=0.25\linewidth]{retrieve/998/k_0.jpg}
    \includegraphics[width=0.25\linewidth]{retrieve/998/k_1.jpg}
    \includegraphics[width=0.25\linewidth]{retrieve/998/k_2.jpg}
    \label{fig:longjump}}
    \caption{Visualization results of video retrieval.The first column is the video frame of query instances. The rightmost three columns are Top-3 nearest retrieval results.}
    \label{fig:RET}
\end{figure}
% \section{Per Class Comparison}
% We measure the per-class classification difference at UCF101, and find that FAME helps baseline enormously to classify motion-sensitive action pairs, such as WallPushups (40.0\% $\uparrow$) vs HandstandPushups (35.7\% $\uparrow$) vs PushUps (33.3\% $\uparrow$). It shows that our method can better discriminate the motion-heavy categories.
% \begin{figure}
%     \centering
%     \includegraphics[width=\linewidth]{Figure/per_Class.png}
%     \caption{Per-class Top-1 accuracy improvement at UCF101.}
%     \label{fig:my_label}
% \end{figure}
\section{More Results on Something-something V2}
We finetune our pretained model on Something-Something V2~\cite{goyal2017something}. We obtain 53.3\% Top-1 accuracy  with R(2+1)D, which beats RSPNet~\cite{chen2021rspnet} under same resolution. 
%%%%%%%%% REFERENCES

{\small
\bibliographystyle{ieee_fullname}
\bibliography{egbib}
}